\pdfoutput=1

\documentclass[11pt,a4paper]{article}
\usepackage{emnlp-ijcnlp-2019}
\usepackage{times}
\usepackage{latexsym}
\usepackage{xcolor}
\usepackage[normalem]{ulem}

\usepackage{booktabs} 
\usepackage{tabularx}
\usepackage{arydshln}
\usepackage{url}
\usepackage{algorithm}
\usepackage{algorithmicx, algpseudocode}
\usepackage{multirow}
\usepackage{multicol}
\usepackage{graphicx}
\usepackage{caption}
\usepackage{titlesec}
\usepackage[section]{placeins}

\titlespacing{\paragraph}{%
  0pt}{
  0.3\baselineskip}{
  1em}%

\newcommand{\Eq}{\E_{\rvz\sim q_{\vphi}(\rvz|\rvx)}}

\newcommand{\qzx}{q_{\vphi}(\rvz|\rvx)}
\newcommand{\pzx}{p_{\vtheta}(\rvz|\rvx)}
\newcommand{\pxz}{p_{\vtheta}(\rvx|\rvz)}

\newcommand{\pz}{p(\rvz)}

\newcommand{\z}{\rvz}
\newcommand{\x}{\rvx}

\usepackage{amsmath,amsfonts,bm}

\def\eqref#1{equation~\ref{#1}}

\def\1{\bm{1}}

\def\rvx{{\mathbf{x}}}

\def\rvz{{\mathbf{z}}}

\def\vtheta{{\bm{\theta}}}
\def\vphi{{\bm{\phi}}}

\DeclareMathAlphabet{\mathsfit}{\encodingdefault}{\sfdefault}{m}{sl}
\SetMathAlphabet{\mathsfit}{bold}{\encodingdefault}{\sfdefault}{bx}{n}

\newcommand{\E}{\mathbb{E}}

\newcommand{\KL}{D_{\mathrm{KL}}}

\definecolor{yycolor}{RGB}{50, 100, 250}

\definecolor{Orange}{RGB}{255, 89, 66}

\definecolor{DarkRed}{RGB}{102, 23, 83}

\aclfinalcopy 

\newcommand\blfootnote[1]{%
  \begingroup
  \renewcommand\thefootnote{}\footnote{#1}%
  \addtocounter{footnote}{-1}%
  \endgroup
}

\title{A Surprisingly Effective Fix for Deep Latent Variable Modeling of Text}
\author{Bohan Li$^{*1}$, Junxian He$^{*1}$, Graham Neubig$^1$, Taylor Berg-Kirkpatrick$^2$, Yiming Yang$^1$ \\
  $^1$Language Technologies Institute, Carnegie Mellon University \\
  $^2$Department of Computer Science and Engineering, University of California San Diego \\
  {\small \texttt{\{bohanl1,junxianh,gneubig,yiming\}@cs.cmu.edu}, \texttt{tberg@eng.ucsd.edu}}}

\date{}

\begin{document}
\maketitle
\begin{abstract}
When trained effectively, the Variational Autoencoder (VAE) is both a powerful language model and an effective representation learning framework. In practice, however, VAEs are trained with the \textit{evidence lower bound} (ELBO) as a surrogate objective to the intractable marginal data likelihood. This approach to training yields unstable results, frequently leading to a disastrous local optimum known as \textit{posterior collapse}. In this paper, we investigate a simple fix for posterior collapse which yields surprisingly effective results. The combination of two known heuristics, previously considered only in isolation, substantially improves held-out likelihood, reconstruction, and latent representation learning when compared with previous state-of-the-art methods. More interestingly, while our experiments demonstrate superiority on these principle evaluations, our method obtains \textit{a worse ELBO}. We use these results to argue that the typical surrogate objective for VAEs may not be sufficient or necessarily appropriate for balancing the goals of representation learning and data distribution modeling.\blfootnote{\hspace{-1.5mm}$^{*}$Equal contribution.}\footnote{Code is available at \url{https://github.com/bohanli/vae-pretraining-encoder}.}

\end{abstract}

\section{Introduction}

Latent variable models attempt to model observed data $\x$ given latent variables $\z$, both for purposes of modeling data distributions $p(\x)$ (e.g.\ language modeling) and learning representations $\z$ for a particular $\x$ (e.g.\ sentence embedding).
Variational Autoencoders (VAEs)~\citep{kingma2013auto} are a powerful framework for learning latent variable models using neural networks.
The generative model of VAEs first samples a latent vector $\z$ from a prior $p(\z)$, then applies a neural decoder $p(\x|\z)$ to produce $\x$ conditioned on the latent code $\z$.
VAEs are trained to maximize the evidence lower bound (ELBO) of the intractable log marginal likelihood:
\begin{equation*}
\label{eq:elbo1}
\Eq[\log\pxz]  - \KL(\qzx \| \pz),
\end{equation*} 
where $\qzx$ represents an approximate posterior distribution (i.e.\ the \textit{encoder} or \textit{inference network}) and $\pxz$ is the generative distribution (i.e.\ the \textit{decoder}). 

However, modeling text with VAEs has proven to be challenging, and is an open research problem~\citep{yang2017improved,xu2018spherical,kim2018semi,dieng2018avoiding,he2018lagging,pelsmaeker2019effective}. When a strong decoder (e.g.\ the LSTM~\citep{hochreiter1997long}) is employed, training often falls into a trivial local optimum where the decoder learns to ignore the latent variable and the encoder fails to encode any information. This phenomenon is referred to as ``posterior collapse''~\citep{bowman2015generating}. 
Existing efforts tackling this problem include
re-weighting the KL loss~\citep{bowman2015generating,kingma2016improved,liu2019cyclical}, changing the model~\citep{yang2017improved,semeniuta2017hybrid,xu2018spherical}, and modifying the training procedure~\citep{he2018lagging}.

After conducting an empirical examination of the state-of-the-art methods (Section~\ref{sec:analysis}), we find that
they have difficulty striking a good balance between language modeling and representation learning. In this paper, we present a practically effective combination of two simple heuristic techniques for improving VAE learning: (1) pretraining the inference network using an autoencoder objective and (2) thresholding the KL term in the ELBO objective (also known as ``free bits''~\citep{kingma2016improved}). The former technique initializes VAE training with an inference network that encodes useful information about $\x$, biasing learning away from local optima where $\x$ is ignored (i.e. posterior collapse). The latter technique modifies the ELBO objective to prevent the KL term from dominating the encoder's role in reconstruction~\citep{alemi2018fixing}, again biasing learning to avoid posterior collapse~\citep{chen2016variational,zhao2017infovae}.

In experiments we find that these two techniques do not perform well in isolation. However, when combined, they substantially outperform all baselines across various metrics that evaluate both language modeling and representation learning capabilities. Finally, we find that our method tends to achieve a superior language modeling results in terms of perplexity but an inferior ELBO value, and we use these results to argue that ELBO is suboptimal for language modeling even though it provides a formal lower bound on log marginal likelihood. Thus, we suggest that future research in this direction should be careful to monitor the gap between ELBO and log marginal likelihood, and reconsider using ELBO as the surrogate, especially for evaluation.

\section{Analysis of Existing Methods}
\label{sec:analysis}

In this section, we first analyze and compare state-of-the-art solutions to posterior collapse\footnote{Here we focus on different training techniques for existing models, not alternatives which require changing the underlying model~\citep{yang2017improved,xu2018spherical}.} to obtain a holistic view of current progress and remaining challenges. Based on these observations, we then propose and evaluate a method that demonstrates substantially improved performance.

\subsection{Evaluation}

In this paper we focus on the standard VAE setting where both the prior $\pz$ and the posterior $\pzx$ are factorized Gaussians.\footnote{~\citet{pelsmaeker2019effective} thoroughly investigate using more complicated priors/posteriors~\citep{rezende2015variational} but find only marginal improvements. } We conduct preliminary experiments on the English Penn Treebank (PTB)~\citep{marcus19building}, a standard dataset for benchmarking VAE that has been used extensively in previous work~\cite{bowman2015generating,xu2018spherical,liu2019cyclical}. We use 32-dimension latent codes (full setup details can be found in Appendix~\ref{sec:appendix:setup}). We evaluate using the following metrics: 
\paragraph{Perplexity (PPL).} 
To compute perplexity, we first estimate the log marginal likelihood with 1000 importance weighted samples~\citep{burda2015importance}. Note that it is inappropriate to estimate PPL with ELBO directly since the gap between ELBO and log marginal likelihood might be large, particularly when the posterior does not collapse. 
\paragraph{Reconstruction loss (Recon).}
Reconstruction loss is equivalent to the negative reconstruction term in ELBO: $-\Eq[\log\pxz]$. It characterizes how well the latent code can be used to recover the input.

\paragraph{Number of active units (AU,~\citet{burda2015importance}). } Active units correspond to the dimensions of $\z$ that covary with observations after the model is trained. More active units usually indicates richer latent representations~\citep{burda2015importance}. Specifically, a dimension is ``active'' when it is sensitive to the change in observations $\rvx$. Here we follow \citep{burda2015importance} and classify a latent dimension $z$ as active if $\textrm{Cov} (\rvx, E_{z \sim q(z | \rvx) } [z])) > 0.01$.

\vspace{+2mm}
In addition to the metrics above, we include KL between prior and posterior approximation, as well as the negative ELBO, for reference -- though we find that these quantities are only partially descriptive of model quality. In Section~\ref{sec:latent-prob} we also evaluate the latent space of learned models with specialized metrics such as reconstruction BLEU and classification accuracy.

\subsection{Baselines}

We experiment with several state-of-the-art techniques to mitigate posterior collapse, including several KL reweighting methods and the recently proposed aggressive training~\citep{he2018lagging}.
\paragraph{KL annealing~\citep{bowman2015generating}. }
KL annealing might be the most common method for reweighting. During annealing, the weight of the KL term is increased from a small value to 1.0 in the beginning of training. 

\paragraph{Cyclic annealing~\citep{liu2019cyclical}. }
Cyclic annealing is another reweighting scheme proposed recently. It changes the weight of the KL term in a cyclic fashion, rather than monotonically increasing the weight.\footnote{We use the default cyclic annealing schedule from \url{https://github.com/haofuml/cyclic_annealing} in our codebase for a fair comparison.}

\paragraph{KL Thresholding / Free Bits (FB) ~\citep{kingma2016improved}. }
FB replaces the KL term in ELBO with a hinge loss term that maxes each component of the original KL with a constant: 
\begin{equation}
\label{eq:freebits}
\sum\nolimits_i \max [\lambda, \KL( q_{\vphi}(z_i|\rvx) \| p(z_i) )]
\end{equation}
Here, $\lambda$ denotes the \textit{target rate}, and $z_i$ denotes the $i^{th}$ dimension in the latent variable $\z$. Using the FB objective causes learning to give up trying to drive down KL for dimensions of $\z$ that are already beneath the target rate. 
\citet{pelsmaeker2019effective} conduct a comprehensive experimental evaluation of related methods and 
conclude that the FB objective is able to achieve comparable or superior performance (in terms of both language modeling and reconstruction) in comparison with other top-performing methods, many of which are substantially more complex. 
We notice that~\citet{pelsmaeker2019effective} experiment with a slightly different version of FB where the threshold is applied to the entire KL term directly, rather than on each dimension's KL separately. We examine both versions here and refer to the single threshold version as ``FBP" and the multiple threshold version (Eq.~\ref{eq:freebits}) as ``FB''. For both FB and FBP, we vary the target rate $\lambda$ and report the setting with the best validation PPL and the setting with the best balance between PPL and reconstruction loss.\footnote{It is subjective to judge ``balance'', thus we also report complete results for different target rates in Appendix~\ref{sec:appendix:detailed_result}.}

\paragraph{Aggressive training~\citep{he2018lagging}. }
 \citet{he2018lagging} observe that when posterior collapse occurs, the inference network often lags behind the generator during training. In contrast with the KL reweighting methods described above, \citet{he2018lagging} propose an aggressive training schedule which iterates between multiple encoder update steps and one decoder update step to mitigate posterior collapse.\footnote{We use the public code at \url{https://github.com/jxhe/vae-lagging-encoder}.} 

\paragraph{Autoencoder (AE). }
We also include an autoencoder\footnote{Here, AE denotes the VAE trained without the KL term.} as a reference for reconstruction loss. 

\begin{table}[!t]
\caption{Results on PTB test set for various baselines.} 
\label{tab:sec2:ptb:anneal_and_fb}
\centering
\scalebox{0.75}{
\begin{tabular}{lrrrrrr}
\toprule
\textbf{Method}& \textbf{PPL}$\downarrow$ & \textbf{Recon}$\downarrow$ & \textbf{AU}$\uparrow$& \textbf{KL}  & \textbf{-ELBO} \\ \midrule
AE & - & 70.36 & 32 & - & -\\\hdashline
VAE & 101.39 & 101.27 & 0 & 0.00 & 101.27 \\
~+ anneal & 101.40 & 101.28 & 0 & 0.00 & 101.28\\
~+ cyclic & 108.97 & 101.85 & 5 & 1.37 & 103.22\\
~+ aggressive & 99.83 & 100.26 & 4 & 0.93 & 101.19\\
~+ FBP ($\lambda=7$) & 102.82 & 95.63 & 4 & 7.05 & 102.67\\
~+ FBP ($\lambda=3$) & 99.62 & 98.52 & 3 & 2.95 & 101.48\\
~+ FB ($\lambda=7$) & 104.06 & 98.97 & 32 & 6.74 & 105.72\\
~+ FB ($\lambda=3$) & 100.50 & 99.94 & 32 & 2.96 & 102.90\\\bottomrule
\end{tabular}
}
\vspace{-10pt}
\end{table}

\vspace{+2mm}
We show the results of these baselines trained on PTB in Table~\ref{tab:sec2:ptb:anneal_and_fb}, where we find that it is difficult to balance language modeling (PPL) and representation learning (Recon and AU) -- the systems with relatively good reconstruction (FBP, $\lambda=7$) or higher AU (FB baselines) have suboptimal PPL, and the best PPL is achieved with sacrifice of Recon and AU. Note that good PPL indicates good LM, but good recon and AU indicates good representation learning. Without both, we do not really have a strong probabilistic model of language that captures latent factors.

We make two observations from the results in Table~\ref{tab:sec2:ptb:anneal_and_fb}. First, reconstruction loss for an AE is substantially better than all VAE methods, which is intuitive since reconstruction is the only goal of training an AE. Second, models with high ELBO do not necessarily have good PPL (e.g.\ VAE+anneal); ELBO is not an ideal surrogate for evaluating language modeling performance. 

\subsection{Autoencoder-based Initialization}
\label{sec:ae-init}
Based on the observations above we hypothesize that VAEs might benefit from initialization with an non-collapsed encoder, trained via an AE objective. Intuitively, if the encoder is providing useful information from the beginning of training, the decoder is more likely to make use of the latent code. 
 In Table~\ref{tab:sec2:ptb:pretrain} we show the results of exploring this hypothesis on PTB. Even with encoder pretraining, we see that posterior collapse occurs immediately after beginning to update both encoder and decoder using the full ELBO objective. This indicates that the gradients of ELBO point towards a collapsed local optimum, even with biased initialization. When pretraining is combined with annealing, PPL improves substantially. 
 However, the pretraining and anneal combination only has 2 active units and has small KL value -- the latent representation is likely unsatisfactory. We speculate that this is because the annealing schedule eventually returns to the full ELBO objective which guides learning towards a (nearly) collapsed latent space. In the next section, we present an alternate approach using the KL thresholding / free bits method. 

\begin{table}[!t]
\caption{Results on PTB test with encoder pretraining.}
\vspace{-7pt}
\label{tab:sec2:ptb:pretrain}
\centering
\resizebox{1.0 \columnwidth}{!}{
\begin{tabular}{lrrrrrr}
\toprule
\textbf{Method}& \textbf{PPL}$\downarrow$ & \textbf{Recon}$\downarrow$ & \textbf{AU}$\uparrow$& \textbf{KL}  & \textbf{-ELBO} \\ \midrule
AE & - & 70.36 & 32 & - & -\\
VAE & 101.39 & 101.27 & 0 & 0.00 & 101.27 \\
~+ pretrain & 102.26 & 101.46 & 0 & 0.00 & 101.46\\
~+ pretrain + anneal & 97.74 & 99.67 & 2 & 1.01 & 100.68\\\bottomrule
\end{tabular}}
\vspace{-15pt}
\end{table}


\subsection{Our Method}
\label{sec:our_method}
In our proposed method, we initialize the inference network with an encoder that is pretrained using an autoencoder objective, as described above. Then, we train the VAE using the FB objective, also described above. We use the original FB which thresholds along each dimension.\footnote{Note that we do not pretrain the decoder and rather initialize it randomly. In our preliminary experiments, pretraining the decoder produced worse PPL compared with encoder-only pretraining. 
} Thus, we combine two approaches so far considered independently: pretraining and KL thresholding. 
In this way, however, the VAE would start with a
large KL and is thus trained with the full ELBO
objective in the initial stage, which prefers the collapsed local optimum as observed in Section~\ref{sec:ae-init}.
To remedy this, we apply an annealing weight to Eq.~\ref{eq:freebits}. We use the simplest linear annealing schedule that increases the weight from 0 to 1 in the first 10 epochs for all of our experiments. This approach can be viewed in connection with KL annealing: we train with zero KL weight until convergence, then reset the decoder and start increasing the KL weight with the KL thresholding objective. 
 Next we conduct comprehensive experiments across different datasets to validate our method.





\section{Experiments}
In this section we work with three text datasets: PTB~\citep{marcus19building}, Yahoo~\citep{yang2017improved}, and a downsampled version of SNLI~\citep{bowman2015large}. We demonstrate the effectiveness of our method through language modeling, text reconstruction, and quality of the learned latent space. Complete experimental setup details can be found in Appendix~\ref{sec:appendix:setup}. 
\label{sec:experiments}
\subsection{Language Modeling}
For language modeling, we only report the results for PTB and Yahoo due to the space limit, but include the SNLI results in Appendix~\ref{sec:appendix:detailed_result}. 
Since we have already shown that FBP outperforms FB on PPL in Section~\ref{sec:analysis} (without pretraining), here we only include FBP as a baseline.
For both FBP and our method we vary the target rate and report the settings that achieve competitive validation PPL.\footnote{Results of all target rates can be found in Appendix~\ref{sec:appendix:detailed_result}.} 



\begin{table}[!t]
\caption{Language modeling results on PTB and Yahoo test set. We bold the lines that represent the best average of  language modeling and reconstruction loss. Cyclic$^{\star}$ is from~\citep{liu2019cyclical}.
} 
\vspace{-7pt}
\label{tab:exp:language_modeling}
\centering
\resizebox{1.0 \columnwidth}{!}{\begin{tabular}{lrrrrrr}
 \toprule
 \textbf{Method} & \textbf{PPL}$\downarrow$ & \textbf{Recon}$\downarrow$ & \textbf{AU}$\uparrow$  & \textbf{KL} & \textbf{-ELBO}\\ \midrule
\multicolumn{6}{c}{\textbf{PTB}} \\
LSTM-LM & 100.47 & - & - & - & -\\
VAE & 101.39 & 101.27 & 0 & 0.00 & 101.27\\
~+ anneal & 101.40 & 101.28 & 0 & 0.00 & 101.28\\
~+ cyclic$^{\star}$ & - & 100.51 & - & 1.96 & 102.46\\
~+ cyclic & 108.97 & 101.85 & 5 & 1.37 & 103.22\\
~+ aggressive & 99.83 & 100.26 & 4 & 0.93 & 101.19\\
~+ FBP ($\lambda=3$) & 99.62 & 98.52 & 3 & 2.95 & 101.48\\
~+ FBP ($\lambda=2$) & 100.96 & 99.37 & 2 & 1.99 & 101.36\\ \hdashline
Ours ($\lambda=8$) &  98.07 & 92.60 & 32 & 10.95 & 103.56\\ 
Ours ($\lambda=6$) &  \bf 96.35 & \bf 94.52 & \bf 32 & 8.15 & 102.67\\ 
Ours ($\lambda=4$) &  96.17 & 96.91 & 32 & 4.99 & 101.90\\ 
\midrule
\multicolumn{6}{c}{\textbf{Yahoo}} \\
LSTM-LM & 60.75 & - & - & - & -\\ 
VAE & 61.52 & 329.10 & 0 & 0.00 &329.10\\
~+ anneal &  61.21 & 328.80 & 0  & 0.00 & 328.80\\
~+ cyclic & 66.93 & 333.80 & 4 & 2.83 & 336.63\\ 
~+ aggressive & 59.77 & 322.70 &	15 &	5.70 & 328.40\\ 
~+ FBP ($\lambda=9$) & 62.59 & 322.91 & 6 & 9.08 & 331.99 \\ 
~+ FBP ($\lambda=7$) & 62.76 & 324.66 & 5 & 7.03 & 331.69\\ 
~+ FBP ($\lambda=5$) & 62.78 & 326.26 & 3 & 5.07 & 331.32 \\ 
~+ FBP ($\lambda=3$) & 62.88 & 328.13 & 2 & 3.06 & 331.19\\  \hdashline
Ours  ($\lambda=6$)  & 59.23 & 317.39 & 32 & 12.09 & 329.48 \\ 
Ours  ($\lambda=8$) & \bf 59.51 & \bf 315.31 & \bf 32  & 15.02& 330.33\\ 
Ours  ($\lambda=9$) & 59.60 & 315.09 &  32  & 15.49& 330.58\\
\bottomrule
\end{tabular}}
\vspace{-15pt}
\end{table}

As shown in Table~\ref{tab:exp:language_modeling}, our method with different target rates is able to consistently outperform all the baselines in terms of PPL. 
Meanwhile, for representation learning, our method beats all the baselines by a large margin on reconstruction loss with all latent units active. 
We also note that our method is not very sensitive to the target rate $\lambda$. 


It is worth noticing that in some cases our method (e.g. $\lambda=8$ in PTB and $\lambda=8,9$ in Yahoo) is able to outperform all the baselines but produce a bad ELBO (actually the worst on PTB). This suggests that ELBO might be a suboptimal surrogate for the log marginal likelihood sometimes, especially when KL is large, where the gap between ELBO and the marginal tends to be large as well.

\subsection{Probing the Latent Space}
\label{sec:latent-prob}
We assess quality of learned latent space with SNLI through several metrics. For our method and FBP, we use the target rate where the best reconstruction loss is achieved while maintaining comparable PPL with aggressive training. \footnote{Basically we try to tie PPL of different models and compare their latent space. Specifically, aggressive training has PPL 32.83, and PPL of the selected model for FBP and our method are 33.07 and 32.88, respectively.}


\paragraph{Text Reconstruction}
We use greedy decoding and compute the BLEU score of the reconstructed sentence with the original one as the reference. The result is shown in Table~\ref{tab:exp:reconstruction}. Unsurprisingly, the autoencoder achieves the highest BLEU score, meanwhile our method beats other VAE baselines.


\begin{table}[!htb]
\vspace{-5pt}
\begin{minipage}{0.5\linewidth}
\caption{\label{tab:exp:reconstruction}Reconstruction}
\vspace{-7pt}
\centering
\scalebox{0.8}{
\begin{tabular}{lr}
\toprule
\textbf{Method} & \textbf{BLEU} \\ \midrule
AE & 60.80 \\
VAE & 1.82 \\
~+ anneal &  2.51 \\
~+ cyclic & 4.39 \\
~+ aggressive & 2.95 \\
~+ FBP ($\lambda=7$) & 8.07 \\\hdashline
Ours ($\lambda=4$)  & \bf 8.62 \\
\bottomrule
\end{tabular}}
\end{minipage}%
\begin{minipage}{0.5\linewidth}
\caption{\label{tab:exp:smoothness}Smoothness}
\vspace{-7pt}
\centering
\scalebox{0.8}{
\begin{tabular}{lrrrr}
\toprule
\textbf{Method} & \multicolumn{3}{c}{\textbf{PCC}} \\ \midrule
AE  & 0.620 \\
VAE  & 0.039\\
~+ anneal  & 0.009\\
~+ cyclic  & 0.482\\
~+ aggressive  & 0.209 \\
~+ FBP ($\lambda=7$)  & 0.242 \\\hdashline
Ours ($\lambda=4$) & \bf 0.683\\
\bottomrule
\end{tabular}}
\end{minipage} 
\vspace{-10pt}
\end{table}

\begin{table}[!t]
\caption{Noisy reconstruction loss ($\downarrow$) on SNLI. \#swap denotes the number of  word swaps.}
\vspace{-7pt}
\label{tab:exp:NoisyReconstruction}
\centering
\scalebox{0.8}{
\begin{tabular}{lrrrr}
\toprule
\textbf{\#swap} &\textbf{1} & \textbf{2} & \textbf{3} & \textbf{4}  \\ \midrule
AE  & \bf 26.05 & 40.46 & 52.77 & 63.07\\
VAE  & 33.10 & 33.11 & 33.11 & 33.12 \\
~+ anneal  & 32.20 & 32.65 & 33.12 & 33.39 \\
~+ cyclic  & 31.83 & 32.87 & 33.73 & 34.38\\
~+ aggressive  & 31.78 & 31.99 & 32.21 & 32.32\\
~+ FBP ($\lambda=7$)  & 29.93 & 32.59 & 34.90 & 36.77\\\hdashline
Ours ($\lambda=4$) & 27.92 &\bf 	29.12 &\bf 	30.03 &\bf 	30.85\\
\bottomrule
\end{tabular}
}
\end{table}

\paragraph{Smoothness of Latent Space}
A major difference between VAE and AE is that VAE can learn a smooth latent space through the regularization from the Gaussian prior. In a smooth latent space, latent codes of similar sentences should be close to each other and vice versa. Therefore, we randomly sample 100k sentence pairs and evaluate the Pearson Correlation Coefficient (PCC) between the $\ell_2$ distances of latent codes and edit word distances. As shown in Table~\ref{tab:exp:smoothness}, our method achieves a much higher PCC compared to the baselines.

\citet{zhao2018adversarially} argue that a smooth latent space  
is beneficial for
reconstructing noisy inputs. We follow their experiments and introduce noise to the input by 
randomly swapping words $k$ times. As shown in Table~\ref{tab:exp:NoisyReconstruction}, while AE achieves the best reconstruction  when the noise is small ($k=1$), its reconstruction deteriorates dramatically when $k>1$, which suggests AE fails to learn a smooth latent space. In contrast, our method outperforms all the baselines by a large margin when $k>1$.


\begin{table}[!h]
\caption{Interpolation between prior samples on SNLI.}
\vspace{-7pt}
\label{tab:exp:interpolation}
\centering
\hspace{-10pt}
\scalebox{0.75}{
\begin{tabular}{l}
\toprule
\textbf{AE} \\
people on their ground and they sit towards each other .  \\
girls riding their cellphones and other people sit near papers .  \\
girls riding in an area while not talk to dishes . \\
person riding in an area while carrying bags and papers .  \\
someone riding in an office , selling a button .  \\
three kid riding in $<$unk$>$ signs a brick advertisement area .  \\\hline
\textbf{Ours}\\
a man with a cane is walking down the street . \\
a man with a cane is walking down the street . \\
a man in a blue shirt is eating food . \\
people are eating food . \\
people walk in a city . \\
people are outside in a city . \\
\bottomrule
\end{tabular}
}
\end{table}


\paragraph{Interpolation} 
As illustrated in \citep{bowman2015generating}, linear interpolation between latent variables is an intuitive way to qualitatively evaluate the smoothness of the latent space. We sample two latent codes $\rvz_0$ and $\rvz_1$ from the prior $\pz$ (Table~\ref{tab:exp:interpolation}) and do linear interpolation between the two with evenly divided intervals.\footnote{More interpolation examples from both prior and posterior are provided in Appendix~\ref{sec:appendix:more_interpolation_examples}.} For each interpolated point, we decode it greedily. Our method is able to generate grammatically plausible and semantically consistent interpolation in both cases. 

\subsection{Classification}
\begin{table}[!t]
\caption{Accuracy on Yelp of unsupervised and supervised classification. Evaluated via accuracy. \#labeled denotes the number of labeled example during training.}
\vspace{-7pt}
\label{tab:exp:classification}
\centering
\scalebox{0.8}{
\begin{tabular}{lrrrrrr}
\toprule
\bf \#labeled& \textbf{0} & \textbf{100} & \textbf{500} & \textbf{1k} & \textbf{2k} & \textbf{10k}  \\ \midrule

AE   & 52.0 & 78.4 & 81.1 & 83.5 & 83.8 & 83.8\\
VAE  & 56.5 & 58.9 & 62.3 & 62.5 & 62.9 & 64.0 \\
~+ anneal & 56.1 & 58.7 & 60.6 & 61.5 & 61.4 & 64.1 \\
~+ cyclic  & 59.3 & 78.1 & 79.8 & 81.1 & 81.7 & 83.1 \\
~+ aggressive  & 63.7 & 65.6 & 68.6 & 72.1 & 76.7 & 79.4 \\
~+ FBP ($\lambda=9$)  & 60.7 & 73.3 & 75.0 & 76.1 & 77.6 & 79.5\\\hdashline
Ours ($\lambda=6$) & \bf 67.2 & \bf 83.8 & \bf 88.3 & \bf 89.1 & \bf 89.5 & \bf 89.5\\
\bottomrule
\end{tabular}
}
\vspace{-15pt}
\end{table}
To further evaluate the quality of the latent codes, we train a Gaussian mixture model (for unsupervised clustering) or a one-layer linear classifier (for supervised classification) on the pretrained latent codes. We work with a downsampled version of Yelp sentiment dataset collected by~\citet{shen2017style}.
We vary the number of labeled data\footnote{Specifically, we train a VAE model on the Yelp dataset to obtain the latent codes, then we we use the latent codes of labeled data to train the classifier.} and the results are shown in Table~\ref{tab:exp:classification}. For FBP and our method, the target rate is selected in terms of the average validation accuracy. Our method consistently yields the best results on all settings -- remarkably, its performance with only 100 labeled samples already surpasses others with 10k labels.

\section{Conclusion}
In this paper, we propose a simple training fix to tackle posterior collapse in VAEs. Extensive experiments demonstrate the effectiveness of our method on both representation learning and language modeling. 

\bibliography{emnlp-ijcnlp-2019}
\bibliographystyle{acl_natbib}

\newpage
\appendix
\onecolumn

\section{Details of Experimental Setup}
\label{sec:appendix:setup}
For SNLI, we randomly downsample a subset of it,
which contains 100K/10K/10K sentences as training/validation/test. For the Yelp sentiment dataset, we also randomly downsample 100K/10K/10K sentences for training/validation/test, respectively.
We use an one-layer LSTM for both the encoder and decoder and a latent vector size of 32. 

We follow \citet{kim2018semi,he2018lagging} and use a single-layer LSTM  the encoder and  decoder in all of our experiments. The sizes of word embeddings and hidden states for different datasets are given in Table~\ref{tab:exp:hidden_size}. We initialize the LSTM parameters with a uniform distribution $\mathcal{U} (-0.01, 0.01)$, and embeddings with another uniform distribution 
$\mathcal{U} (-0.1, 0.1)$.

Just as in \citet{kim2018semi, he2018lagging}, for the decoder, we use dropout of 0.5 on both the input word embedding and the last dense before logits. During training, we use the SGD optimizer wihout momentum. Initialized with 0.5, the learning rate is decayed by $\times 0.5$ with a patience of 2 if the validation loss has not improved in the past 2 epochs. The maximum number of epochs is 100, but the training will stop early after 5 learning rate decays. For our VAE + anneal baseline, we use the simplest linear annealing schedule that increases the weight from 0 to 1 in the first 10 epochs, just the same as in our method (stated in Section~\ref{sec:our_method}).

\begin{table}[h]
\caption{The sizes of word embeddings and hidden states for PTB, SNLI and Yahoo.}
\label{tab:exp:hidden_size}
\centering
\footnotesize
\setlength\tabcolsep{2.5pt}
\begin{tabular}{lrrr}
\toprule
 & PTB & SNLI & Yahoo \\ \midrule
Word Embedding Size & 256 & 128 & 512 \\
Hidden Size of Encoder & 256 & 512 & 1024 \\
Hidden Size of Decoder & 256 & 512 & 1024 \\
\bottomrule
\end{tabular}
\end{table}

\section{Copying Behaviour}
We check to make sure that our model is not simply learning to copy sentences from the training set. We test for copying behaviour on the PTB dataset. Specifically, we sampled 300 sentences from the prior and retrieved their nearest neighbors in the training set. The average edit distance between the samples and their nearest neighbors from our method is 14.11 (the average training sentence length is 22.10), versus 13.68 from the collapsed VAE. This means that there is no obvious copying behaviour when KL grows in our method. 

\section{Additional Results of Language Modeling}
\label{sec:appendix:detailed_result}
In Table~\ref{tab:appendix:detailed_result}, we provide a more detailed version of the language modeling results on PTB, SNLI and Yahoo. The full results of all the target rate trial for both FBP and our method are included. In addition to the metrics in Table~\ref{tab:exp:language_modeling}, we also report NLL, Mutual Information $I_q$ between $\rvz$ and $\rvx$ under $\qzx$ (MI)\footnote{For the estimation of $I_q$, we follow the same method as in \citet{he2018lagging}}, and the perplexity computed by ELBO (ELBO PPL).

\begin{table*}[!h]
\caption{Additional results of language modeling on PTB and SNLI.}
\label{tab:appendix:detailed_result}
\centering
\scalebox{0.7}{
\begin{tabular}{l|lrrrrrrrrr}
 \toprule
\textbf{Dataset} & \textbf{Method} & \textbf{NLL}$\downarrow$ & \textbf{PPL}$\downarrow$  & \textbf{Recon}$\downarrow$ & \textbf{MI}$\uparrow$ & \textbf{AU}$\uparrow$  & \textbf{-ELBO} & \textbf{ELBO PPL} &\textbf{KL}\\ \midrule
\multirow{4}{*}{PTB} 
&LSTM-LM & 101.04	& 100.47 &- &- &- & -&- & -\\
&AE &  &  & 70.36 & 8.22 & 32 & - & - & -\\ 
&VAE& 101.23 & 101.39 & 101.27 & 0.01 & 0 & 101.27 & 101.58 & 0.00 \\ 
&~+ anneal & 101.24 & 101.40 & 101.28 & 0.00 & 0 & 101.28 & 101.62 & 0.00 \\ 
&~+ cyclic (reported) & - & - & 100.51 & - & - & 102.46 & 107.25 & 1.96\\ 
&~+ cyclic  & 102.81 & 108.97 & 101.85 & 1.27 & 5 & 103.22 & 111.03 & 1.37\\ 
&~+ cyclic (SGD) & 102.06 & 105.28 & 102.14 & 0.00 & 0 & 102.14 & 105.67 & 0.00\\ 
&~+ aggressive  & 100.89 & 99.83 & 100.26 & 0.83 & 4 & 101.19 & 101.17 & 0.93\\ 
&~+ pretrained enc & 101.42 & 102.26 & 101.46 & 0.0 & 0 & 101.46 & 102.45 & 0.00\\
&~~~+ anneal enc & 100.43 & 97.74 & 99.67 & 0.97 & 2 & 100.68 & 98.88 & 1.01\\
&~+ FBP ($\lambda=9$)& 101.95 & 104.73 & 94.66 & 7.22 & 6 & 103.59 & 112.88 & 8.93\\
&~+ FBP ($\lambda=8$)& 101.64 & 103.30 & 95.39 & 7.11 & 6 & 103.43 & 112.09 & 8.04\\
&~+ FBP ($\lambda=7$)& 101.54 & 102.82 & 95.63 & 6.52 & 4 & 102.67 & 108.26 & 7.05\\
&~+ FBP ($\lambda=6$)& 101.68 & 103.47 & 96.66 & 5.76 & 10 & 102.67 & 108.27 & 6.01\\
&~+ FBP ($\lambda=5$)& 101.24 & 101.42 & 97.12 & 4.80 & 4 & 102.21 & 106.01 & 5.10\\
&~+ FBP ($\lambda=4$)& 101.49 & 102.58 & 97.84 & 3.86 & 4 & 101.86 & 104.31 & 4.01\\
&~+ FBP ($\lambda=3$)& 100.85 & 99.62 & 98.52 & 2.86 & 3 & 101.48 & 102.52 & 2.95\\
&~+ FBP ($\lambda=2$)& 101.14 & 100.96 & 99.37 & 1.91 & 2 & 101.36 & 101.98 & 1.99\\
& IWAE ($k=10$)& 100.86 & 99.69 & 100.89 & 0.05 & 0 & 100.89 & 99.81 & 0.00 \\ 
&~+ pretrained enc & 100.92 & 99.95 & 100.96 & 0.01 & 0 & 100.96 & 100.15 & 0.00\\ 
&Ours ($\lambda=9$) & 101.09 & 100.71 & 92.00 & 7.78 & 32 & 104.43 & 117.32 & 12.44\\
&Ours ($\lambda=8$) & 100.51 & 98.07 & 92.60 & 7.49 & 32 & 103.56 & 112.72 & 10.95\\
&Ours ($\lambda=7$) & 101.06 & 100.60 & 93.25 & 7.46 & 32 & 103.74 & 113.68 & 10.49 \\
&Ours ($\lambda=6$) & 100.12 & 96.35 & 94.52 & 6.30 & 32 & 102.67 & 108.25 & 8.15 \\ 
&Ours ($\lambda=5$) & 100.23 & 96.86 & 95.87 & 5.31 & 32 & 102.41 & 106.97 & 6.54\\ 
&Ours ($\lambda=4$) & 100.08 & 96.17 & 96.91 & 4.08 & 32 & 101.90 & 104.51 & 4.99 \\ 
&Ours ($\lambda=3$) & 100.21 & 96.75 & 97.71 & 3.19 & 32 & 101.56 & 102.90 & 3.85\\
&Ours ($\lambda=2$) & 100.41 & 97.65 & 98.73 & 2.21 & 32 & 101.38 & 102.07 & 2.65\\
\midrule
\multirow{4}{*}{SNLI} 
&  LSTM-LM & 32.97	& 21.44 &- &- &- & -&- & -\\
&AE& - & - & 8.68 & 9.18 & 32 & - & - & -\\ 
&VAE & 33.09 & 21.67 & 33.08 & 0.03 & 1 & 33.12 & 21.73 & 0.04\\ 
&~+ anneal & 33.01 & 21.50 & 31.66 & 1.45 & 2 & 33.07 & 21.63 & 1.42\\ 
&~+ cyclic & 34.04 & 23.67 & 30.69 & 3.60 & 5 & 34.32 & 24.29 & 3.63\\ 
&~+ cyclic (SGD) & 33.07 & 21.62  & 30.89 & 2.33 & 4 & 33.25 & 21.99 & 2.36 \\
&~+ aggressive & 32.83 & 21.16 & 31.53 & 1.38 & 5 & 32.95 & 21.39 & 1.42\\ 
&~+ FBP ($\lambda=9$)& 33.28 & 22.05 & 25.26 & 8.06 & 6 & 34.25 & 24.13 & 8.99\\ 
&~+ FBP ($\lambda=8$)& 33.26 & 22.02 & 26.07 & 7.35 & 7 & 34.08 & 23.75 & 8.01\\ 
&~+ FBP ($\lambda=7$)& 33.07 & 21.62 & 26.65 & 6.76 & 6 & 33.78 & 23.11 & 7.14\\ 
&~+ FBP ($\lambda=6$)& 33.09 & 21.68 & 27.54 & 5.95 & 5 & 33.59 & 22.71 & 6.06\\ 
&~+ FBP ($\lambda=5$)& 33.04 & 21.58 & 28.38 & 4.95 & 6 & 33.49 & 22.48 & 5.10\\ 
&~+ FBP ($\lambda=4$)& 33.04 & 21.57 & 29.25 & 4.06 & 4 & 33.36 & 22.22 & 4.11\\ 
&~+ FBP ($\lambda=3$)& 33.04 & 21.56 & 30.19 & 3.00 & 4 & 33.31 & 22.11 & 3.11\\ 
&~+ FBP ($\lambda=2$)& 32.99 & 21.46 & 31.04 & 2.11 & 3 & 33.16 & 21.80 & 2.12\\ 
&Ours ($\lambda=9$) & 33.42 & 22.33 & 22.30 & 8.80 & 32 & 35.70 & 27.61 & 13.40 \\
&Ours ($\lambda=8$) & 33.47 & 22.45 & 22.65 & 8.76 & 32 & 35.62 & 27.40 & 12.96\\
&Ours ($\lambda=7$) & 33.25 & 22.00 & 23.36 & 8.48 & 32 & 35.11 & 26.14 & 11.75\\
&Ours ($\lambda=6$) & 33.17 & 21.84 & 24.06 & 8.24 & 32 & 34.83 & 25.47 & 10.77\\
&Ours ($\lambda=5$) & 33.07 & 21.64 & 24.94 & 7.71 & 32 & 34.47 & 24.63 & 9.53\\ 
&Ours ($\lambda=4$) & 32.88 & 21.24 & 26.52 & 6.65 & 32 & 34.11 & 23.83 & 7.60\\ 
&Ours ($\lambda=3$) & 32.87 & 21.23 & 28.02 & 5.25 & 32 & 33.87 & 23.31 & 5.86\\ 
&Ours ($\lambda=2$) & 32.79 & 21.07 & 29.75 & 3.32 & 32 & 33.42 & 22.35 & 3.67\\
\bottomrule
\end{tabular}}
\end{table*}

\newpage
\begin{table*}[!h]
\caption{Additional results of language modeling on Yahoo.}
\label{tab:appendix:detailed_result:yahoo}
\centering
\scalebox{0.7}{
\begin{tabular}{l|lrrrrrrrrr}
 \toprule
\textbf{Dataset} & \textbf{Method} & \textbf{NLL}$\downarrow$ & \textbf{PPL}$\downarrow$  & \textbf{Recon}$\downarrow$ & \textbf{MI}$\uparrow$ & \textbf{AU}$\uparrow$  & \textbf{-ELBO} & \textbf{ELBO PPL} &\textbf{KL}\\ \midrule
\multirow{4}{*}{Yahoo} 
& LSTM-LM & 328.00 & 60.75 &- &- &- & -&- & -\\
&AE & - & - & 278.76 & 9.26 & 32 & - & - & -\\ 
&VAE & 329.00 & 61.52 &329.10 &0.00 & 0 & 329.10 & 61.59 & 0.0\\
&~+ anneal & 328.60 &61.21 & 328.80 & 0.00 & 0 & 328.80 & 61.36& 0.0 \\
&~+ cyclic & 335.74 & 66.93 & 333.80 & 2.77 & 4 & 336.63 & 67.69 & 2.83 \\ 
&~+ cyclic (SGD)& 332.48 & 64.26 & 332.65 & 0.00 & 1 & 332.68 & 64.42 & 0.03 \\
&~+ aggressive & 326.70 & 59.77 & 322.70 & 2.9 & 15 & 328.40 & 61.06 & 5.70\\
&~+ FBP ($\lambda=9$)& 330.38 & 62.59 & 322.91 & 8.21 & 6 & 331.99 & 63.86 & 9.08\\ 
&~+ FBP ($\lambda=8$)& 330.80 & 62.92 & 324.03 & 7.54 & 6 & 332.09 & 63.94 & 8.05\\ 
&~+ FBP ($\lambda=7$)& 330.60 & 62.76 & 324.66 & 6.76 & 5 & 331.69 & 63.62 & 7.03\\ 
&~+ FBP ($\lambda=6$)& 331.05 & 63.12 & 325.87 & 5.94 & 5 & 332.01 & 63.88 & 6.13\\ 
&~+ FBP ($\lambda=5$)& 330.62 & 62.78 & 326.26 & 5.00 & 3 & 331.32 & 63.33 & 5.07\\ 
&~+ FBP ($\lambda=4$)& 331.06 & 63.13 & 327.55 & 4.00 & 3 & 331.66 & 63.60 & 4.11\\ 
&~+ FBP ($\lambda=3$)& 330.75 & 62.88 & 328.13 & 2.99 & 2 & 331.19 & 63.23 & 3.06\\ 
&~+ FBP ($\lambda=2$)& 331.30 & 63.32 & 329.60 & 1.98 & 1 & 331.63 & 63.58 & 2.04\\ 
&Ours  ($\lambda=2$) & 326.34 & 59.50 & 322.55 & 5.35 & 32 & 328.51 & 61.14 & 5.96\\ 
&Ours  ($\lambda=3$) & 326.12 & 59.34 & 321.29 & 6.41 & 32 & 328.73 & 61.31 & 7.44\\
&Ours  ($\lambda=4$) & 326.01 & 59.26 & 319.49 & 7.58 & 32 & 329.03 & 61.54 & 9.54\\
&Ours  ($\lambda=5$) & 326.04 & 59.28 & 318.55 & 8.08 & 32 & 329.31 & 61.76 & 10.76\\
&Ours  ($\lambda=6$) & 325.97 & 59.23 & 317.39 & 8.51 & 32 & 329.48 & 61.89 & 12.09\\ 
&Ours  ($\lambda=7$) &  326.08 & 59.31 & 316.42 & 8.78 & 32 & 329.76 & 62.10 & 13.34\\
&Ours  ($\lambda=8$) & 326.35 & 59.51 & 315.31 & 8.99 & 32 & 330.33 & 62.55 & 15.02\\ 
&Ours  ($\lambda=9$) & 326.47 & 59.60 & 315.09 & 9.03 & 32 & 330.58 & 62.75 & 15.49\\
\bottomrule
\end{tabular}}
\end{table*}

\section{Additional Qualitative Examples}
\label{sec:appendix:more_interpolation_examples}

\subsection{Interpolation between Prior Samples}
We randomly sample 10 pair of source and target latent code from the standard Gaussian prior and do linear interpolation. For the sampled and interpolated latent codes, we do greedy decoding. The results are shown in Table~\ref{tab:appendix:prior_interpolation}.

\begin{table}[!h]
\caption{Interpolation between prior samples on SNLI}
\label{tab:appendix:prior_interpolation}.
\centering
\scalebox{0.7}{
\begin{tabular}{ll}
\toprule
EXAMPLE 1 & EXAMPLE 6\\ \midrule
the kids are playing hide and seek in the classroom   & a man is jumping off a rock into the air .  \\ 
the girl is about to play the drums   & a man is sitting on a bench with a red umbrella .  \\ 
the girl is about to play in the sandbox   & a man is sitting on a bench with a red umbrella .  \\ 
the girl is about to play in the sandbox   & a man is sitting on a bench with a red umbrella .  \\ 
the girls are watching tv in the classroom .   & a woman is sitting on a bench in front of a building .  \\ 
the girl is eating cake in the kitchen .   & a young man is sitting on a bench outside .  \\ 
the women are watching tv in the bar .   & a young man is taking pictures of a building .  \\ 
the women are eating lunch .   & a large group of people are taking pictures of a building .  \\ 
the women are eating lunch .   & a large group of people are taking pictures in the street .  \\ 
the women are eating dinner .   & a large group of people are taking pictures in the street .  \\ 
a woman is eating in a restaurant .   & a large group of people are taking pictures in the street .  \\ \midrule
EXAMPLE 2 & EXAMPLE 7\\ \midrule
a man with a cane is walking down the street .   & two men are swimming in the ocean .  \\ 
a man with a cane is walking down the street .   & two men are swimming in the ocean .  \\ 
a man with a cane is walking down the street .   & two men are on the beach .  \\ 
a man with a cane is walking down a sidewalk .   & two people are at the beach .  \\ 
a man in a blue shirt is eating food .   & two people are at the beach .  \\ 
man in a hat and jeans is walking down a sidewalk .   & the man is at the park .  \\ 
people are eating food .   & a man is at the beach .  \\ 
people walk through a city street .   & a man is at the beach .  \\ 
people walk in a city .   & a man is at the beach .  \\ 
people walk in a city .   & a man is at the beach .  \\ 
people are outside in a city .   & a man is taking pictures of the ocean .  \\ \midrule
EXAMPLE 3 & EXAMPLE 8\\ \midrule
the man is going to the bathroom .   & two people in bathing suits are in a park .  \\ 
the man is going to the bathroom .   & two people in bathing suits are in a park .  \\ 
the man is going to the bathroom .   & two people in blue shirts are in a field .  \\ 
the man is going to the bathroom .   & two people in blue shirts are in a field .  \\ 
the man is playing music in the living room .   & two people in a field are playing soccer .  \\ 
the man is playing with his dog .   & two people in a field of flowers .  \\ 
the man is playing with a cat .   & two people are at a beach .  \\ 
the man is playing with a ball .   & a man in a blue shirt is looking at a camera .  \\ 
the man is playing with a ball .   & a man in a blue shirt is playing basketball .  \\ 
the man is playing with a ball .   & a man in a blue shirt is playing basketball .  \\ 
the man is playing with a ball .   & a man sits at a carnival .  \\ \midrule
EXAMPLE 4 & EXAMPLE 9\\ \midrule
a person is about to get a picture taken   & two women are outside .  \\ 
a person is about to get a picture taken   & two women are outside .  \\ 
a person is about to get a picture taken   & two women are outside .  \\ 
a person is waiting for a friend to come to work   & two women are sitting outside .  \\ 
a person is waiting for a friend to come   & a couple is sitting outside .  \\ 
a person is trying to find a speech   & a couple is sitting outside .  \\ 
the people are playing monopoly   & a couple is sitting in a car .  \\ 
there are people playing monopoly   & a man is reading a newspaper in a park .  \\ 
there are people performing   & a man is reading a book in a park .  \\ 
there are people performing surgery   & a man is carrying a bag of food .  \\ 
there are no people in this picture .   & a man is carrying a bag of food .  \\ \midrule
EXAMPLE 5 & EXAMPLE 10\\ \midrule
a girl sits on a bench in front of a large crowd .   & the man is climbing the mountain .  \\ 
a girl sits on a bench in front of a large crowd .   & the man is making a noise .  \\ 
a man sits on a bench in front of a large crowd .   & the man is making a noise .  \\ 
a man sits on a bench in front of a large crowd .   & the man is making a noise .  \\ 
a woman wearing a blue shirt is walking on the beach .   & the people are enjoying the sunshine .  \\ 
a woman wearing a blue shirt is walking on the beach .   & the people are watching a movie .  \\ 
the woman is wearing a blue shirt .   & the women are eating a meal .  \\ 
the woman is wearing a blue shirt .   & the men are watching a movie .  \\ 
the woman is wearing a blue shirt .   & the men are watching a movie .  \\ 
the woman is wearing a blue shirt .   & two women sit in a circle together .  \\ 
the woman is moving her legs .   & two women sit in a circle together .  \\
\bottomrule
\end{tabular}
}
\end{table}

\subsection{Interpolation between Posterior Samples}
We randomly sample 10 pairs of source and target input sentences from the test set of SNLI. For each input sentence, we randomly sample a latent code from the approximated posterior $q(\rvz|\rvx)$. Then we linearly interpolate between each pair of sampled source/target latent code. For the sampled and interpolated latent codes, we do greedy decoding. The results are shown in Table~\ref{tab:appendix:post_interpolation}.

\begin{table}[!h]
\caption{Interpolation between posterior samples on SNLI.}
\label{tab:appendix:post_interpolation}
\centering
\scalebox{0.6}{
\begin{tabular}{lll}
\toprule
~& EXAMPLE 1 & EXAMPLE 6\\ \midrule
SOURCE INPUT & a child is eating with utensils . & a middle-aged man with long , curly red-hair wearing a dark vest , \\ 
~ &  & shirt and pants is holding a microphone in front of a black backdrop .\\ \hdashline
TARGET INPUT & a youth wearing a blue and red jersey and yellow helmet is  & people are doing $<$unk$>$\\ 
~ & crouching in a football position & \\ \hdashline
POSTERIOR-SAMPLED SOURCE & a little girl is eating at a restaurant . & the two men are wearing jeans and a blue shirt , and a woman \\ 
&&are holding a rope .\\ \hdashline
INTERPOLATION & a little girl is eating at a restaurant . & the two men are wearing jeans and a blue shirt .\\ 
 & a little girl is eating at a table . & the two men are wearing jeans and a blue shirt .\\ 
 & a little girl is playing with a toy . & the two men are wearing white shirts and are playing a sport .\\ 
 & a little girl is playing with a ball . & the people are trying to find a cure for cancer .\\ 
 & a little girl is wearing a pink shirt . & the people are playing soccer\\ 
 & a little girl is wearing a pink shirt and holding a popsicle . & two people are playing baseball\\ 
 & a man is wearing a blue shirt and a hat . & two people are playing baseball\\ 
 & a man is wearing a blue shirt and a hat . & two people are playing baseball\\ 
 & a man is wearing a black shirt and black pants & people are playing baseball\\ \hdashline
POSTERIOR-SAMPLED TARGET & a man wearing a black shirt and black pants is standing on a sidewalk & people are playing baseball\\ \midrule
 & EXAMPLE 2 & EXAMPLE 7\\ \midrule
SOURCE INPUT & the men are feeling competetive . & both men are wearing similar colors .\\ \hdashline
TARGET INPUT & a young woman is sitting in a field . & a huge animal surrounded\\ \hdashline
POSTERIOR-SAMPLED SOURCE & a man is climbing a tree . & a man is cutting a cake .\\ \hdashline
INTERPOLATION & a man is climbing a tree . & a man is cutting a cake .\\ 
 & a man is eating a pizza . & a man is painting a portrait of a woman .\\ 
 & a man is wearing a blue shirt . & a man is painting a portrait of a woman .\\ 
 & a man is wearing a blue shirt . & a man with a beard is playing guitar .\\ 
 & a man is wearing a blue shirt . & a young boy with a blue shirt .\\ 
 & a man is standing in front of a large crowd . & a little girl with a pink shirt .\\ 
 & a woman is standing in front of a large crowd . & a tall human with a shirt\\ 
 & a woman is standing in front of a large crowd . & a tall human with a shirt\\ 
 & a woman is standing in front of a large crowd . & a tall human with a shirt\\ \hdashline
POSTERIOR-SAMPLED TARGET & a woman is standing in front of a large crowd . & a tall human looking\\ \midrule
 & EXAMPLE 3 & EXAMPLE 8\\ \midrule
SOURCE INPUT & the animals are near the water . & a crowded city street in asia .\\ \hdashline
TARGET INPUT & a truck is going to tow an illegally parked white volkswagon . & a guy is in front of a business . a tree with flowers is in front .\\ \hdashline
POSTERIOR-SAMPLED SOURCE & two boys are at a beach . & a bunch of people are in a store .\\ \hdashline
INTERPOLATION & two boys are at a beach . & a bunch of people are in a park .\\ 
 & two men are looking at a man in a wheelchair . & a few people are in a park .\\ 
 & the children are at the beach . & a few people are in a park .\\ 
 & the children are looking at the sky . & a few people are in a park .\\ 
 & a woman is looking at a man in a wheelchair . & a few people are in front of a building .\\ 
 & a woman is looking at a man in a wheelchair . & a lady is in a store with a man .\\ 
 & a woman is looking at a map . & a lady in a blue shirt is looking at a man in a blue shirt .\\ 
 & a woman is waiting for a bus to come out of the road . & a lady in a blue shirt is looking at a man in a blue shirt .\\ 
 & a woman is waiting for a bus to come out of the city . & a lady in a blue shirt is looking at a man in a blue shirt .\\ \hdashline
POSTERIOR-SAMPLED TARGET & a woman is waiting for a bus . & a lady in a blue shirt and black pants is playing in a fountain with \\
&& a small child in the background .\\\midrule
 & EXAMPLE 4 & EXAMPLE 9\\ \midrule
SOURCE INPUT & one young child in a swimsuit jumping off a blue inflatable slide & a woman with 5 small children .\\  
& with water .  &\\\hdashline
TARGET INPUT & a girl swings from a rope swing in front . & a man works on $<$unk$>$ a circuit as he monitors the progress \\ 
&&on a tablet device .\\\hdashline
POSTERIOR-SAMPLED SOURCE & a young man in a blue shirt and black pants is standing  & man with blue shirt and blue shirt is playing basketball .\\
&by a large rock formation .&\\\hdashline
INTERPOLATION & a young girl in a pink shirt and blue shorts is jumping into a pool . & man with blue shirt and blue shirt is playing basketball .\\ 
 & a woman in a pink shirt and black shorts is jumping into a pool . & man with blue shirt and blue shirt is playing basketball .\\ 
 & a woman in a pink shirt and black shorts is playing a game of soccer . & man in blue shirt with a blue shirt on his head .\\ 
 & a woman in a pink shirt and black shorts is playing a game of soccer . & a man with a hat is playing with a ball .\\ 
 & a woman with a red shirt and a black shirt is sitting in a chair . & a man in a blue shirt is looking at a plant .\\ 
 & a woman with a red shirt and a black shirt is sitting in a chair . & a man in a blue shirt is sitting on a bench with a shovel .\\ 
 & a woman with a red shirt and a black shirt is sitting in a chair . & a man in a blue shirt is sitting on a bench with a shovel .\\ 
 & a woman with a red shirt and a black shirt is looking at a camera . & a man is on a skateboard in front of a building .\\ 
 & a woman watches a man play a game of soccer . & a man is on a skateboard in front of a building with graffiti on it .\\  \hdashline
POSTERIOR-SAMPLED TARGET & a woman watches a man play a guitar in front of a crowd . & a man works on a project on a stove .\\ \midrule
 & EXAMPLE 5 & EXAMPLE 10\\ \midrule
SOURCE INPUT & a girl with glasses next red white and blue flags . & a man walking across a bridge near a steak restaurant .\\  \hdashline
TARGET INPUT & three greyhounds are taking a walk with their owner . & a woman in a white shirt and shorts is playing a red guitar .\\  \hdashline
POSTERIOR-SAMPLED SOURCE & the girl is drinking milk with the camera . & people in a park\\  \hdashline
INTERPOLATION & the girl is drinking milk with the camera . & a woman in a blue shirt is eating a sandwich .\\ 
 & the girl is drinking milk with her hands . & a woman in a blue shirt is playing a game of tennis .\\ 
 & the girl is drinking water with a bucket . & a woman is playing a game of tennis .\\ 
 & the girl is using a camera . & a woman is playing a game of tennis .\\ 
 & two girls are outside with a blue umbrella . & a man is playing a guitar .\\ 
 & two girls are outside with a blue umbrella . & a man is playing a guitar .\\ 
 & two girls are outside with a dog . & a man is playing a guitar .\\ 
 & two girls are taking a picture of a tree . & a man is playing a guitar .\\ 
 & two guys are on a bench . & a man is playing a guitar .\\  \hdashline
POSTERIOR-SAMPLED TARGET & two guys are on a boat . & a man is playing a guitar .\\
\bottomrule
\end{tabular}
}
\end{table}
\end{document}